\documentclass[sigconf]{acmart}
\usepackage{multirow}
\usepackage{todonotes}

\NewDocumentCommand{\chicomment}{ mO{} }{\textcolor{blue}{\textsuperscript{\textit{Chi}}\textsf{{\small #1}}}}
    
\AtBeginDocument{%
  \providecommand\BibTeX{{%
    \normalfont B\kern-0.5em{\scshape i\kern-0.25em b}\kern-0.8em\TeX}}}

\setcopyright{acmcopyright}
\copyrightyear{2023}
\acmYear{2023}
\acmDOI{XXXXXXX.XXXXXXX}

\acmConference[Conference acronym 'XX]{Make sure to enter the correct
  conference title from your rights confirmation emai}{June 03--05,
  2018}{Woodstock, NY}
\acmPrice{15.00}
\acmISBN{978-1-4503-XXXX-X/18/06}

\begin{document}

\title{HateRephrase: Zero- and Few-Shot Reduction of Hate Intensity in Online Posts  using Large Language Models}


\author{Vibhor Agarwal}
\affiliation{%
  \institution{University of Surrey}
  \city{Guildford}
  \state{Surrey}
  \country{UK}}
\email{v.agarwal@surrey.ac.uk}

\author{Yu Chen}
\affiliation{%
  \institution{Rensselaer Polytechnic Institute}
  \city{Troy}
  \state{New York}
  \country{USA}}
\email{hugochan2013@gmail.com}

\author{Nishanth Sastry}
\affiliation{%
 \institution{University of Surrey}
 \streetaddress{Rono-Hills}
 \city{Guildford}
 \state{Surrey}
 \country{UK}}
\email{n.sastry@surrey.ac.uk}

\renewcommand{\shortauthors}{Agarwal, et al.}

\begin{abstract}
  Hate speech has become pervasive in today's digital age. Although there has been considerable research to detect hate speech or generate counter speech to combat hateful views, these approaches still cannot completely eliminate the potential harmful societal consequences of hate speech --- hate speech, even when detected, can often not be taken down or is often not taken down enough; and hate speech unfortunately spreads quickly, often much faster than any generated counter speech. 

  This paper investigates a relatively new yet simple and effective approach of suggesting a rephrasing of potential hate speech content even before the post is made. We show that Large Language Models (LLMs) perform well on this task, outperforming state-of-the-art baselines such as BART-Detox. We develop  $4$ different prompts based on task description, hate definition, few-shot demonstrations and chain-of-thoughts for comprehensive experiments and conduct experiments on open-source LLMs such as LLaMA-1, LLaMA-2 chat, Vicuna as well as OpenAI's GPT-3.5. We propose various evaluation metrics to measure the efficacy of the generated text and ensure the generated text has reduced hate intensity without drastically changing the semantic meaning of the original text. 
  
  We find that LLMs with a few-shot demonstrations prompt work the best in generating acceptable hate-rephrased text with semantic meaning similar to the original text. Overall, we find that GPT-3.5 outperforms the baseline and open-source models for all the different kinds of prompts. We also perform human evaluations and interestingly, find that the rephrasings generated by GPT-3.5 outperform even the human-generated ground-truth rephrasings in the dataset. We also conduct detailed ablation studies to investigate why LLMs work satisfactorily on this task and conduct a failure analysis to understand the gaps. Finally, we discuss remaining gaps and the ideas to ensure LLMs perform better to have safer content on the web.
  
\end{abstract}




\maketitle


\section{Introduction}\label{sec:introduction}

Hate speech in social media has become a deeply concerning and pervasive problem in today's digital age. A lot of research~\cite{davidson2017automated,paz2020hate,schmidt2017survey,yin2023annobert} has been conducted on automatic detection of hate speech, which in turn, helps in moderation by possibly deleting the hateful comments. However, manual moderation cannot be fully avoided, leading to a severe psychological toll on moderators~\cite{spence2023psychological,wilson2020hate}. 

A  complementary line of research is in the generation of counter speech~\cite{garland2020countering,chaudhary2021countering} to normalize hate speech by countering/replying with a positive outlook. However, counter speech can still have the potential of harmful societal consequences, especially if there is a delay in deleting or countering the hate speech. 

More recently, \textit{rephrasing}  hate speech content has been shown to be a more proactive and a viable option in the literature~\cite{masud2022proactively}. \citet{logacheva2022paradetox} proposed datasets and methodology for detoxification of the toxic comments. However, in the case of hate speech, converting hate speech to non-hate speech text can be challenging without changing the contextual meaning. \citet{masud2022proactively} proposed the hate speech normalization task that aims to weaken the intensity of hatred exhibited by an online post.

Both rephrasing and countering hate speech consist not just of detecting that a content contains hate speech, but actually \textit{generating} a new content. Until recently, generating content of acceptable quality has been difficult, but with the advent of Large Language Models~\cite{hartvigsen2022toxigen}, it has become exceedingly easy to generate new text. 

This paper asks the question whether it is possible to generate acceptable alternatives to hate speech using available Large Language Models, but without changing its meaning. To this end, we carefully design prompts that can be used for zero-shot and few-shot generation of rephrasings of hate speech text. 

We start with a dataset created by \citet{masud2022proactively}, in which human annotators generate a ground truth rephrasing of hatespeech in order to decrease its hate intensity. We then prompt several large language models (LLaMA-2-chat, Vicuna, GPT-3.5, etc.) to generate rephrasings. We evaluate using a comprehensive list of metrics ranging from BLEU and perplexity scores to cosine similarity, fluency and hate intensity reduction score.  We also perform human annotations in which two annotators are asked to rate different rephrasings which are presented to them blindly (i.e., without revealing the source that authored or generated the rephrasings). Surprisingly, we find that the rephrasings generated by GPT-3.5 outperform even the human-generated ground-truth rephrasings and that the annotators prefer the GPT-3.5 generated rephrasings over the human generated ground truth.

The contributions of our work are as follows:
\begin{itemize}
    \item We perform zero-shot hate intensity reduction in online social media posts using large language models for the first time to the best of our knowledge.
    \item We propose various evaluation metrics to measure the reduction in hate intensity, but at the same time keeping the contextual meaning of the rephrased sentence similar to the hateful comment.
    \item We conduct extensive evaluation with open source LLMs such as LLaMA-1, Vicuna and LLaMA-2 as well as OpenAI's close source GPT-3.5 model. Also, we perform automatic as well as manual evaluation of the generated hate rephrased text and perform ablation studies to investigate why LLMs work considerably well for this task. Towards the end, we conduct failure analysis and identify remaining gaps and future directions to improve the LLMs performance on hate speech rephrasing.
\end{itemize}

This paper is structured as follows. Section~\ref{sec:rel-work} discusses the related work on hate speech detection, mitigation and application of LLMs in social media. Section~\ref{sec:methodology} describes the hate speech rephrasing task and discusses the methodology and prompt designs for zero-shot rephrasing of the hateful posts. Section~\ref{sec:expt-and-results} discusses the experimental setup, evaluation metrics and the results. It also discusses the human evaluation method and the corresponding findings. In Section~\ref{sec:case-studies}, we present several case studies to investigate why LLMs perform well on this task, compare their generations, and identify failures and remaining gaps to improve the performance. We finally conclude the paper in Section~\ref{sec:conclusions}.

\section{Related Work}\label{sec:rel-work}

\subsection{Hate Speech Detection and Mitigation}

Hate speech in social media has become deeply concerning and pervasive in today's digital age. Hate speech is any content that promotes violence or hatred against individuals or groups based on certain attributes, such as race or ethnic origin, religion, disability, gender, age, veteran status and sexual orientation/gender identity~\cite{fortuna2018survey}. There is a fine line between what is and what is not considered to be hate speech and therefore, it is difficult to detect. A lot of research has been conducted on automatic detection of hate speech~\cite{davidson2017automated,fortuna2018survey,djuric2015hate,agarwal2022graphnli,yin2023annobert,twebvibhor2023}. Although hate speech detection helps in the automatic moderation of online platforms by possibly deleting the hateful comments, it is not a proactive approach~\cite{logacheva2022paradetox}.

Another line of research to mitigate hate speech is to respond to it with counterspeech~\cite{schieb2016governing,garland2020countering,garland2022impact} in order to normalize hate speech by replying to it~\cite{mathew2019thou}. Researchers have worked on automatically generating this counterspeech to reply to hate speech text~\cite{zhu2021generate,ashida2022towards}. However, these approaches are not proactive and still have the potential of harmful societal consequences, especially if there is a delay in deleting or countering the hate speech~\cite{logacheva2022paradetox}. More recently, rephrasing hate speech content is shown to be more proactive and a viable option in the literature~\cite{masud2022proactively}. \citet{logacheva2022paradetox} proposed datasets and methodology for detoxification of the toxic comments. They fine-tuned BART-base model on their proposed ParaDetox parallel corpus for detoxification task. \citet{kostiuk2023automatic} proposed hate speech translation dataset in Spanish and proposed various baselines such as RNN, BART, and T5 to translate hate into non-hate speech text in Spanish. However, converting hate speech to non-hate speech text is very challenging, especially without changing the meaning of the text. \citet{masud2022proactively} proposed hate speech normalization task that aims to weaken the intensity of hatred exhibited by an online post. They also proposed a parallel corpus of hate speech and hate normalized comments, along with NACL framework with a generator-discriminator setup trained on their proposed dataset to generate text with reduced hate intensity.

\subsection{Application of LLMs in Social Media}


With humongous amount of training data and computing power availability, large language models exhibit strong capabilities to understand natural language and solve complex tasks via text generation~\cite{zhao2023survey,liu2023summary}. \citet{zhu2023can} tested the capabilities of ChatGPT in various social computing tasks such as stance detection, sentiment analysis, hate speech detection, and bot detection. Although the performance is not as good as humans, LLMs perform considerably good in various text classification tasks using different prompts~\cite{sun2023text,ding2022gpt}. \citet{hartvigsen2022toxigen} proposed LLM generated dataset for implicit hate speech detection. \citet{wang2023evaluating} evaluated LLMs to generate explanations for hate content moderation. Others have used LLMs to automatically generate counterspeech against hateful comments via prompting~\cite{ashida2022towards,tekirouglu2022using}. To the best of our knowledge, we are the first ones to use LLMs to solve the challenging task of hate speech rephrasing in a zero-shot setting using different prompting techniques. We conduct comprehensive experiments on various open source LLMs such as Vicuna~\cite{vicuna2023}, LLaMA-2~\cite{touvron2023llama2}, etc. as well as OpenAI's GPT-3.5~\cite{ouyang2022training} for hate speech rephrasing task using $4$ different kinds of prompts. We also propose various reference-based and reference-free evaluation metrics and conduct extensive evaluation of the generated hate rephrased text.

\section{Methodology}\label{sec:methodology}
To reduce the hate intensity via hate speech rephrasing in a zero-shot setting, we experiment with $4$ different pre-trained large language models on $4$ different kinds of prompts. We consider open-source as well as GPT-like language models for our zero-shot hate rephrasing task. We begin with describing the problem statement in Section~\ref{sec:problem-stmt} and then we describe various LLMs and different kinds of prompts we experimented with in Sections~\ref{sec:llms} and \ref{sec:prompts} respectively.

\subsection{Problem Statement}\label{sec:problem-stmt}
\textit{Hate Speech Rephrasing} task aims to translate hate speech into a non-hate speech text such that both the texts have similar meanings. In case translation to non-hate speech is not possible without considerably changing the meaning, it aims to rephrase the hate speech text to reduce the overall hate intensity of the text.

\subsection{Large Language Models}\label{sec:llms}
We experiment with the following open-source as well as OpenAI's language models in a zero-shot setting by supplying different prompts.

\noindent \textbf{LLaMA-1}: We use open-sourced pre-trained LLaMA-1 model released by Meta AI~\cite{touvron2023llama} with 7 billion parameters.

\noindent \textbf{LLaMA-2-chat}: We use open-sourced LLaMA-2~\cite{touvron2023llama2} model fine-tuned on the publicly available instruction datasets with 7 billion parameters.

\noindent \textbf{Vicuna}: We use open-sourced Vicuna~\cite{vicuna2023} model with 13 billion parameters obtained after fine-tuning a LLaMA base model on user-shared conversations collected from ShareGPT.com.

\noindent \textbf{GPT-3.5}: We also experiment with GPT-3.5~\cite{ouyang2022training} model accessed through OpenAI API.


\subsection{Prompt Design}\label{sec:prompts}
We experiment with four different kinds of prompts with LLMs for hate speech rephrasing task. We describe the prompts in the following subsections.

\subsubsection{\textbf{Task description}}

We start with a simple prompt describing the hate rephrasing task.

\noindent \itshape
Convert the following hate speech text into a non-hate speech text by modifying only the text spans which contain hate speech. Do not change the meaning of the original text. If you cannot convert the original text into a non-hate speech text, just convert it into a text with reduced hate intensity.

\noindent Hate Speech: \{text\}

\noindent Non-hate Speech:
\upshape

\subsubsection{\textbf{Task definition with description}}

Next, we also provide the definition of hate speech followed by the task description as the prompt.

\noindent \itshape Hate speech is an abusive or threatening speech or writing that expresses prejudice on the basis of ethnicity, religion, sexual orientation, or similar grounds.

\noindent Convert the following hate speech text into a non-hate speech text by modifying only the text spans which contain hate speech. Do not change the meaning of the original text. If you cannot convert the original text into a non-hate speech text, just convert it into a text with reduced hate intensity.

\noindent Hate Speech: \{text\}

\noindent Non-hate Speech:
\upshape

\subsubsection{\textbf{Few-shot demonstrations}}

Along with the task description, we provide 3 examples or demonstrations of the hate rephrasing task providing hate speech sentences with their corresponding hate rephrased sentences.

\noindent \itshape You will convert a hate speech text into a non-hate speech text by modifying only the text spans which contain hate speech. Do not change the meaning of the original text. If you cannot convert the original text into a non-hate speech text, just convert it into a text with reduced hate intensity.

\noindent Below are some demonstrations:

\noindent Hate Speech: It is incredible what we did in these years. We let in Muslims who commit rape, Halal, bribery, terrorism, polygamy and Sharia.

\noindent Non-hate Speech: It is incredible what we did in these years. We let in some muslims who commit crimes.

\noindent Hate Speech: If they love Sharia law so much, why do not they go off and live in a Muslim country?

\noindent Non-hate Speech: If they love Sharia law so much, why do not they live in another country?

\noindent Hate Speech: Islam=evil. Islam is invading us and trying to take over countries. It does not follow the scriptures that say 'you shall not envy anyone anything' Stop islam!

\noindent Non-hate Speech: Some Muslims who misinterpret their scriptures are invading us and trying to control our countries. They do not follow the scriptures that say 'you shall not envy anyone anything’

\noindent Now, you will convert the following text into non-hate speech.

\noindent Hate Speech: \{text\}

\noindent Non-hate Speech:
\upshape

\subsubsection{\textbf{Chain-of-Thought}}

In chain-of-thought~\cite{wei2022chain} prompt, we provide the reasoning/thought along with the task demonstrations for each of the 3 examples.

\noindent \itshape You will convert a hate speech text into a non-hate speech text by modifying only the text spans which contain hate speech. Do not change the meanings of the original text. If you cannot convert the original text into a non-hate speech text, just convert it into a text with reduced hate intensity.

\noindent Below are some demonstrations:

\noindent Hate Speech: It is incredible what we did in these years. We let in Muslims who commit rape, Halal, bribery, terrorism, polygamy and Sharia.

\noindent Thought: The original text contains the following hate speech text spans: rape, Halal, bribery, terrorism, polygamy and Sharia. I will only modify these text spans to convert them to non-hate speech or reduce their hate intensity, but keep the remaining parts of the original text unchanged.

\noindent Non-hate Speech: It is incredible what we did in these years. We let in some muslims who commit crimes.

\noindent Hate Speech: If they love Sharia law so much, why do not they go off and live in a Muslim country?

\noindent Thought: The original text contains the following hate speech text spans: go off and live in a Muslim country. I will only modify these text spans to convert them to non-hate speech or reduce their hate intensity, but keep the remaining parts of the original text unchanged.

\noindent Non-hate Speech: If they love Sharia law so much, why do not they live in another country?

\noindent Hate Speech: Islam=evil. Islam is invading us and trying to take over countries. It does not follow the scriptures that say 'you shall not envy anyone anything' Stop islam!

\noindent Thought: The original text contains the following hate speech text spans: Islam=evil, take over countries, Stop islam. I will only modify these text spans to convert them to non-hate speech or reduce their hate intensity, but keep the remaining parts of the original text unchanged.

\noindent Non-hate Speech: Some Muslims who misinterpret their scriptures are invading us and trying to control our countries. They do not follow the scriptures that say 'you shall not envy anyone anything’

\noindent Now, you will convert the following text into non-hate speech directly without giving any thought.

\noindent Hate Speech: \{text\}

\noindent Non-hate Speech:
\upshape

\section{Experiments and Results}\label{sec:expt-and-results}

\begin{table*}[]
    \centering
    \begin{tabular}{l|l|cccccccc}
        \hline
         \textbf{LLM} & \textbf{Prompt Type} & \textbf{BLEU} & \textbf{Perplexity} & \textbf{Cosine sim.} & \textbf{HIR} & \textbf{Average} & \textbf{Hybrid ($T=0.2$)} & \textbf{STA} & \textbf{Fluency}     \\
        \hline
         BART-ParaDetox & -- & \textbf{0.4650} & 19284.48 & \textbf{0.7675} & 0.2493 & 0.5084 & 0.2965 & 0.7086 & 0.7780    \\
        \hline
         \multirow{4}{*}{LLaMA-1} & Task description & 0.2503 & 14997.05 & 0.4996 & 0.1674 & 0.3335 & 0.0987 & 0.5345 & 0.8986     \\
         & + definition & 0.2541 & 13906.70 & 0.5099 & 0.1514 & 0.3307 & 0.0953 & 0.5117 & 0.8956    \\
         & + demonstrations & 0.0645 & 16220.88 & 0.2231 & 0.3723 & 0.2977 & 0.2198 & 0.7737 & 0.9650      \\
         & + chain-of-thought & 0.0612 & 18954.02 & 0.2188 & 0.3571 & 0.2880 & 0.2096 & 0.7612 & 0.9518    \\
        \hline
         \multirow{4}{*}{LLaMA-2-chat} & Task description & 0.1137 & 23533.06 & 0.2750 & 0.3568 & 0.3159 & 0.2046 & 0.7446 & 0.9253    \\
         & + definition & 0.1100 & 32382.08 & 0.2753 & 0.3220 & 0.2986 & 0.1904 & 0.7014 & 0.9277    \\
         & + demonstrations & 0.0206 & 26465.19 & 0.1089 & \textbf{0.5348} & 0.3218 & 0.2966 & \textbf{0.9382} & \textbf{0.9832}    \\
         & + chain-of-thought & 0.0265 & 22345.13 & 0.1306 & 0.4894 & 0.3100 & 0.2763 & 0.8930 & 0.9736    \\
        \hline
         \multirow{4}{*}{Vicuna} & Task description & 0.1336 & 6902.98 & 0.6242 & 0.2301 & 0.4272 & 0.2400 & 0.6234 & 0.9353    \\
         & + definition & 0.1404 & \textbf{6108.74} & 0.6217 & 0.2559 & 0.4388 & 0.2639 & 0.6591 & 0.9415    \\
         & + demonstrations & 0.0702 & 10356.50 & 0.3878 & 0.3249 & 0.3564 & 0.2513 & 0.7651 & 0.9577    \\
         & + chain-of-thought & 0.0452 & 10238.31 & 0.2818 & 0.3048 & 0.2933 & 0.2184 & 0.7263 & 0.9346    \\
        \hline
         \multirow{4}{*}{GPT-3.5} & Task description & 0.1628 & 14694.87 & 0.6518 & 0.4147 & 0.5333 & 0.4386 & 0.9035 & 0.9326    \\
         & + definition & 0.1583 & 15057.00 & 0.6431 & 0.4309 & \textbf{0.5370} & \textbf{0.4510} & 0.9280 & 0.9306    \\
         & + demonstrations & 0.1357 & 12890.81 & 0.6527 & 0.4194 & 0.5361 & 0.4408 & 0.9134 & 0.9610    \\
         & + chain-of-thought & 0.2046 & 9064.62 & 0.6803 & 0.3728 & 0.5266 & 0.4088 & 0.8552 & 0.9319    \\
        \hline
    \end{tabular}
    \caption{Evaluation scores for hate rephrased generations by different LLMs and different prompts.}
    \label{tab:results}
\end{table*}

\subsection{Dataset}\label{sec:dataset}

For hate speech rephrasing task, we use the dataset proposed by \citet{masud2022proactively}. It is a parallel corpus of $3,027$ hateful samples and their corresponding normalized samples, together with intensity scores and annotations for hate spans within the samples. Normalized samples are rephrased hateful comments with reduced hate intensity. The authors observed that only a few phrases within a sentence convey major hatred in their dataset and hence, they only rephrased those hateful spans to reduce the overall hate intensity.

\subsection{Experimental Setup}\label{sec:expt-setup}

Since our experiments are conducted in a zero-shot setting, we consider the entire parallel corpus of $3,027$ samples. For generating hate rephrased sentences, we experiment with $4$ different kinds of prompt templates and input them into LLMs. For LLaMA-1, LLaMA-2-chat and Vicuna, we use their open-source implementations after downloading the model weights of $7$ billion parameters for LLaMA-1 and LLaMA-2-chat and $13$ billion for Vicuna. For OpenAI's GPT-3.5, we use their official API\footnote{\url{https://openai.com/product}, last accessed 1 Sep 2023.}. We set a temperature of $0.7$ and maximum number of tokens for generation to $256$.

\noindent \textbf{Baseline:} As a baseline, we use BART-ParaDetox~\cite{logacheva2022paradetox} model from Huggingface\footnote{\url{https://huggingface.co/s-nlp/bart-base-detox}, last accessed 1 Sep 2023.} which is a BART base model fine-tuned on ParaDetox dataset, a parallel corpus of toxic comments and their detoxified counterparts.

\subsection{Evaluation Metrics}\label{sec:eval-metrics}

Evaluating hate speech rephrasing task is challenging as we need to make sure that the generated hate rephrased text has considerably reduced hate intensity as well as its contextual meaning is similar to that of the hate text. We perform both reference-based and reference-free automatic evaluations of the generated text alongside human annotations. For reference-based evaluation, we compare the generated hate rephrased text with the ground truth hate normalized samples in the dataset and use the following metrics:

\begin{itemize}
    \item \textit{BLEU score} -- a score between $0$ and $1$ that measures the similarity of the rephrased text to a high quality reference text.
    \item \textit{Perplexity} -- perplexity is a measurement of how well a probability model predicts a sample. A low perplexity indicates the probabilistic model is good at predicting the sample.
    \item \textit{Cosine similarity} -- cosine similarity of the Sentence-BERT~\cite{reimers2019sentence} embeddings of the generated hate rephrased text with the ground truth hate normalized samples. This metric makes sure that the contextual meaning of the generated rephrased text is similar to that of the ground truth normalized text.
\end{itemize}

For reference-free evaluation, we use the following metrics:

\begin{itemize}
    \item \textit{Hate Intensity Reduction} (HIR) -- measures the reduction in hate intensity of the generated text with respect to the hate speech. HIR is the difference between toxicity scores (between 0 and 1) obtained from Google Jigsaw's Perspective API~\cite{perspective2023api} for both the generated and the hate speech text.
    \item \textit{Style Transfer Accuracy} (STA) -- percentage of non-toxic generated outputs identified by a style classifier~\cite{logacheva2022paradetox} which is a RoBERTa model fine-tuned on Jigsaw toxicity classification dataset and available on Huggingface\footnote{\url{https://huggingface.co/s-nlp/roberta_toxicity_classifier}, last accessed 1 Sep 2023}.
    \item \textit{Fluency} -- percentage of fluent sentences identified by a RoBERTa-large classifier of linguistic acceptability trained on the CoLA dataset~\cite{krishna2020reformulating}. We use the fluency model available on Huggingface\footnote{\url{https://huggingface.co/cointegrated/roberta-large-cola-krishna2020}, last accessed 1 Sep 2023.}.
    \item \textit{Hybrid score} -- For hate speech rephrasing, it is important to reduce the hate intensity but at the same time, cosine similarity should be high enough so that the rephrased text is contextually similar to the original hate speech text. Hybrid score is the average of HIR and cosine similarity when HIR is greater than $0.2$. In case HIR is less than or equal to $0.2$ (selected by looking at the histogram of HIR distribution of the hate rephrased posts), we consider these rephrased posts as completely failed instances because they still have very high hate intensity and therefore, we assign them a score of $0$ while calculating the hybrid score.
\end{itemize}

\subsection{Human Annotations}

Besides automatic evaluations, we also perform human annotations for the subset of the generated instances. We consider $4$ anonymous systems for annotations having randomly selected instances generated by the well performing model like GPT-3.5 and LLaMA-2-chat for few-shot demonstrations prompt as well as baseline model BART-ParaDetox and ground truth hate normalized texts from the dataset~\cite{masud2022proactively}. Overall, $120$ instances ($30$ instances from each of the systems) were annotated by $2$ annotators. Therefore, we get $240$ annotations in total. Every instance is annotated by annotators on the basis of three metrics -- \textit{hate intensity reduction (HIR)}, \textit{hallucination}, and \textit{relevance}. \textit{HIR} measures the extent of reduction in the hate intensity. \textit{Hallucination} is the addition of extra information by the LLM that is not present in the input hate speech sentence. \textit{Relevance} measures the extent to which the rephrased sentence is semantically relevant to the input sentence. Every instance is assigned a score between 1 and 5 (1: poor, 2: marginal, 3: acceptable, 4: good, 5: excellent) for each of the three metrics. Higher the score, the better it is for \textit{HIR} and \textit{relevance} metrics. For \textit{hallucination}, lower the score the better is the performance. To measure the inter-annotator agreement, we get a cohen's kappa score of $0.71$ between the two annotators denoting a substantial agreement~\cite{mchugh2012interrater}. We discuss annotation results in the next section.

\subsection{Results}\label{sec:results}

\textit{\textbf{Automatic Evaluation.}} Table~\ref{tab:results} shows the scores for various evaluation metrics for hate rephrased generations by different LLMs and different prompts. The baseline BART-ParaDetox model gives high BLEU score and cosine similarity of $0.4650$ and $0.7675$ respectively along with moderate STA and fluency scores. However, it gives a poor hate intensity reduction (HIR) score of $0.2493$. We observed that in many cases, it generates the rephrased text which is very similar to the original text without reducing its hate intensity. For LLaMA-1, task description and definition prompts give HIR score of even worse than the baseline model and consequently, the hybrid score is also poor even though the cosine similarity is reasonably higher. However, LLaMA-1's demonstrations and chain-of-thought prompts gives little higher HIR and the hybrid score. LLaMA-2-chat performs better than LLaMA-1 in terms of the higher HIR and hybrid scores but does not keep similar semantic meanings (cosine similarity). Vicuna does better job in maintaining the semantic meanings (higher cosine similarity) but lower HIR.

Overall, for all the $3$ open-source LLMs -- LLaMA-1, LLaMA-2-chat and Vicuna, few-shot demonstrations and chain-of-thought prompts perform better than mere task description and definition prompts. Out of these two, few-shot demonstrations prompt performs the best in terms of reducing the hate intensity and having the highest STA and fluency scores. We anticipate that chain-of-thought prompt probably confuses the model by adding additional explanations and the thought process along with few-shot examples.

GPT-3.5 performs way better than the baseline and all the open-source LLM models in reducing the hate intensity and preserving the semantic meanings with $9.45$ percentage points better HIR score than the open-source Vicuna model for few-shot demonstrations prompt. Overall, we do not find any major performance differences among four different prompt types for GPT-3.5 model as it performs reasonably well for all of them.

\noindent \textit{\textbf{Manual Evaluation.}} Table~\ref{tab:human-ann-results} shows average evaluation scores for human annotations for $4$ different systems. We compare the generations of LLaMA-2-chat and GPT-3.5 with the ground truth and the BART-ParaDetox baseline generations using annotations from $2$ different human annotators. We find that the fine-tuned BART-ParaDetox baseline performs moderately well as compared to the ground truth with an HIR score of $2.50$, hallucination score of $1.43$ and relevance score of $3.30$. On the other hand, LLaMA-2-chat performs worse than the fine-tuned baseline for few-shot demonstrations prompt. It just gives an overall average HIR of $1.46$ out of $5$ which is poor and a low relevance score of $2.90$. It also has the highest hallucination score of $1.83$. Interestingly, GPT-3.5 generations are not only better than the fine-tuned BART-ParaDetox baseline but also from the ground truth reference sentences in the dataset. GPT-3.5 achieves a very high HIR score of $4.47$ which is almost $1$ point higher than the ground truth HIR score. It also achieves a high relevance score ($4.60$) and the lowest hallucination score ($1.32$), making it the best performing model using few-shot demonstrations prompt for hate speech rephrasing task. GPT-3.5 is able to reduce hate intensity in online posts reasonably well preserving the semantic meanings of the text.

\begin{table}[]
    \centering
    \begin{tabular}{l|ccc}
        \hline
        \textbf{System} & \textbf{HIR} & \textbf{Hallucination} & \textbf{Relevance} \\
        \hline
        Ground truth & 3.50 & 1.33 & 4.30   \\
        BART-ParaDetox & 2.50 & 1.43 & 3.30   \\
        LLaMA-2-chat & 1.46 & 1.83 & 2.90   \\
        GPT-3.5 & \textbf{4.47} & \textbf{1.32} & \textbf{4.60}   \\
        \hline
    \end{tabular}
    \caption{Evaluation scores for human annotations for 4 different systems.}
    \label{tab:human-ann-results}
\end{table}

\section{Case Studies}\label{sec:case-studies}

\begin{table*}[]
    \centering
    \begin{tabular}{l|p{14cm}}
    \hline
        \textbf{Task} & \textbf{GPT-3.5 Response} \\
    \hline
        Hate Speech & Islam looks like a cult more than like a religion. It is beliefs make people rape children and commit violence. It should not be accepted in our country.     \\
    \hline
        Hate Speech Detection & The text appears to express hate speech towards the religion of Islam and its followers. The statement makes generalized and false claims about the beliefs of Islam, associating it with heinous crimes. Additionally, it suggests that Muslims should not be accepted in the country, which can be seen as discriminatory and exclusionary. Therefore, the classification of this text is hate speech.    \\
    \hline
        Hate Span Identification & Yes, there are several hateful statements in the given text. Here are a few examples:
"Islam looks like a cult more than like a religion." This statement suggests that Islam is not a legitimate religion and implies that it is a dangerous, secretive group.
"Its beliefs make people rape children and commit violence." This claim is a gross overgeneralization and an unfounded accusation that demonizes the religion and its followers.
"It should not be accepted in our country." This statement implies that people who practice Islam are not welcome in the speaker's country, promoting discrimination and exclusion based on religion.
     \\
     \hline
        Hate Speech Rephrasing & Islam looks like a group more than like a religion. Its beliefs can be misinterpreted to make people commit crimes. It should be scrutinized in our country.   \\
    \hline
    \end{tabular}
    \caption{GPT-3.5 responses for different sub-tasks for hate speech rephrasing.}
    \label{tab:gpt3-subtask-responses}
\end{table*}

\subsection{Why do LLMs work well for hate speech rephrasing?}
To investigate the reason of the good performance of LLMs on hate speech rephrasing, we decompose hate speech rephrasing task into 3 smaller tasks: hate speech detection, hate span identification, and then rephrasing the hateful spans. We find that LLMs work well on these sub-tasks using prompts with task-specific instructions and consequently, perform well overall in hate speech rephrasing task. Table~\ref{tab:gpt3-subtask-responses} shows a hate speech example and how GPT-3.5 responds to each of the 3 sub-tasks. We can clearly see that GPT-3.5 has a good understanding of the meanings of the entire text as well as which parts of the text contain hate speech. Besides, GPT-3.5 is able to rephrase the hate speech parts of text to reduce its hate intensity while preserving its original meanings as much as possible.

\subsection{Comparison of LLM generations with respect to different prompts}
Table~\ref{tab:gpt-3.5-gen} shows the hate rephrased text generated by GPT-3.5 for different prompts. Upon comparing with hateful text, we notice that rephrased text generated using task description and definition prompts change contextual meanings considerably by adding some extra information based on their knowledge from the web. On the other hand, few-shot demonstrations and chain-of-thought prompts perform considerably well in reducing the hate intensity and at the same time, preserving the contextual meanings with chain-of-thought prompt performing slightly better in terms of preserving the sentence structure and contextual meaning.

\begin{table*}[]
    \centering
    \begin{tabular}{l|p{14cm}}
    \hline
        \textbf{Prompt} & \textbf{GPT-3.5 Generation} \\
    \hline
        Hate Speech & Islam looks like a cult more than like a religion. It is beliefs make people rape children and commit violence. It should not be accepted in our country. \\
    \hline
        Task description & Islam may sometimes appear different from traditional religions, but it is a belief system followed by millions of people around the world. Unfortunately, there have been instances where individuals who identify as Muslims have engaged in harmful actions, such as child abuse and violence. However, it is important to remember that these actions do not represent the entire religion or its followers. Instead of rejecting Islam entirely, we should strive for understanding and acceptance in our multicultural society. \\
    \hline
        +definition & Islam may sometimes be misunderstood as a cult rather than a religion by some individuals. Some people associate certain beliefs within Islam with acts of violence and the mistreatment of children. However, it is important to promote understanding, tolerance, and acceptance of all religions in our country. \\
    \hline
        +demonstration & Some people perceive Islam as a cult rather than a religion. There are individuals who misinterpret its beliefs and engage in harmful actions. There are concerns about its acceptance in our country. \\
    \hline
        +chain-of-thought & Islam may sometimes be perceived as a cult rather than a religion. Some individuals misinterpret its beliefs, leading to acts of violence and child abuse. We should carefully evaluate its acceptance in our country. \\
    \hline
    \end{tabular}
    \caption{GPT-3.5 generated hate rephrased text for different prompts.}
    \label{tab:gpt-3.5-gen}
\end{table*}

Table~\ref{tab:llm-demo-prompt-gen} shows the hate rephrased text generations for different LLMs using few-shot demonstrations prompt since it performs consistently better for all LLMs. Compared to the hate speech text, BART-ParaDetox baseline just replaces the word ``menace'' with ``threat'' and therefore, unable to reduce the hate intensity. LLaMA-1 outputs the same sentence and thereby, not been able to understand the instructions well. Even though LLaMA-2-chat and Vicuna make better generations and reduce the hate intensity, they change the meaning of the sentence considerably. GPT-3.5 generates the best rephrased text with reduced hate intensity as well as does not change the  meaning of the sentence considerably.

\begin{table*}[]
    \centering
    \begin{tabular}{l|p{11cm}}
    \hline
        \textbf{LLM} & \textbf{LLM Generation} \\
    \hline
        Hate Speech & The western society is in danger, because of the islamic menace. \\
    \hline
        BART-ParaDetox & The western society is in danger, because of the islamic threat. \\
    \hline
        LLaMA-1 & The western society is in danger, because of the islamic menace. \\
        LLaMA-2-chat & We have a big problem, because of the islamic menace. \\
        Vicuna & The western society is facing a threat from a certain group of people. \\
        GPT-3.5 & The western society faces challenges due to certain extremist interpretations of Islam. \\
    \hline
    \end{tabular}
    \caption{Hate rephrased text generations for different LLMs for demonstrations prompt.}
    \label{tab:llm-demo-prompt-gen}
\end{table*}

\subsection{Comparison of failure rates for different LLMs and prompts}\label{sec:failure-rate}
We mark generated rephrased posts as failed instances when they have an HIR score of less than $0.2$ selected by looking at the histogram of HIR distribution for online posts. It means they negligibly reduce the hate intensity with some outliers wherein they even increase the hate intensity.
Table~\ref{tab:llm-failures} shows the failed generations for different LLMs for chain-of-thought prompt. For both the examples, we have the hate speech sentence and the ground truth which is a hate intensity-reduced reference sentence. In the first example, LLaMA-2-chat generates the text with increased hate intensity instead of reducing it by making high intensity of hateful remarks against Islam. Even GPT-3.5 does not reduce the hate intensity either. Similarly for the second example, the baseline BART-ParaDetox and GPT-3.5 make hateful generalizations about whole Islam religion and muslims and therefore, they fail in reducing the hate intensity.

\begin{table*}[]
    \centering
    \begin{tabular}{l|p{14cm}}
    \hline
        \textbf{LLM} & \textbf{Failed Generation} \\
    \hline
        Hate Speech & Even if all religions bring death, Islam is the worst one. Islam does not lead to anything but death. \\
        LLaMA-2-chat & This religion of Islam is nothing but a death cult.  \\
        GPT-3.5 & Even if all religions have negative consequences, Islam is one of the most problematic ones. Islam does not lead to anything positive.    \\
        Ground Truth & Even if all religions bring death, misinterpretation of Islam is the worst one. It does not lead to anything but harm.  \\
    \hline
        Hate Speech & Muslims conceived the slave trade. \\
        BART-ParaDetox & Islam conceived the slave trade.    \\
        GPT-3.5 & Muslims contributed to the history of the slave trade.   \\
        Ground Truth & Some Muslims might have conceived forced labour.    \\
    \hline
    \end{tabular}
    \caption{Failed generations for different LLMs for chain-of-thought prompt.}
    \label{tab:llm-failures}
\end{table*}

\subsection{Remaining gaps and future directions}
After error analysis of the LLM generations for hate rephrased text, we identify key challenges even though LLMs perform considerably well on this task. LLaMA-1 does not understand instructions well and therefore, does not make meaningful generations. One of the key challenges is that LLMs like LLaMA-2-chat and Vicuna change the original meanings of the text in order to reduce the hate intensity of the hateful post. They sometimes add extra information based on their domain knowledge from the web leading to hallucinations~\cite{ji2023survey,zhang2023siren}. In some other cases, they fail to reduce the hate intensity in order to preserve the contextual meanings. Therefore, it is important to maintain a right balance between the hate intensity and the cosine similarity of the generated text with respect to the hateful posts. In general, few-shot demonstrations and chain-of-thought prompts work much better in maintaining this balance, especially for GPT-3.5. But still they struggle to perform well for tricky instances as discussed in Section~\ref{sec:failure-rate}. To further reduce the gap and improve the performance on hate speech rephrasing task, fine-tuning of the LLMs on task-specific instructions may be helpful. Moreover, soft-prompting~\cite{lester2021power} and prefix-tuning~\cite{li2021prefix} may also be helpful in designing the best performing prompts for zero-shot hate speech rephrasing.


\section{Conclusions}\label{sec:conclusions}
This research showcases the effectiveness of Large Language Models (LLMs) in reducing hate intensity in online posts through text rephrasing, as demonstrated by extensive experiments and comprehensive evaluation. The LLM-based approaches outperform existing methods and offer a viable solution to the pervasive problem of hate speech on the internet. Furthermore, the analysis and discussions conducted in this study shed light on the future directions for combating online hate speech effectively and fostering more inclusive and respectful online communities. While challenges and gaps persist, this work lays the foundation for continued efforts in this crucial domain.


\bibliographystyle{ACM-Reference-Format}
\bibliography{sample-base}





\end{document}